\documentclass[runningheads]{llncs}
\usepackage{graphicx}
\usepackage{amsmath,amssymb} 
\usepackage{color}
\usepackage[width=122mm,left=12mm,paperwidth=146mm,height=193mm,top=12mm,paperheight=217mm]{geometry}
\begin{document}
\pagestyle{headings}
\mainmatter
\def\ECCV16SubNumber{***}  

\title{Skin Lesion Analysis Towards Melanoma Detection via End-to-end Deep Learning of Convolutional Neural Networks} 

\titlerunning{ISIC2018 submission}

\authorrunning{ISIC2018 submission}

\author{Katherine M. Li and Evelyn C. Li}
\institute{ }

\maketitle

\begin{abstract}
This article presents the design, experiments and results of our solution submitted to the 2018 ISIC challenge: Skin Lesion Analysis Towards Melanoma Detection.
We design a pipeline using state-of-the-art Convolutional Neural Network (CNN) models for a Lesion Boundary Segmentation task and a Lesion Diagnosis task.

\end{abstract}

\section{Introduction}
Recently, state-of-the-art research in Artificial Intelligence (AI) has rapidly transitioned from research laboratories
into production \cite{Stoica:EECS-2017-159}.
This development is driven by a number of forces in the industry:
1) widespread availability of a large amount of data, in the form of images (e.g., ImageNet), audio, video, text and user logs;
2) significant increase in affordable computation power (e.g., NVidia GeForce GTX 1080Ti cards, which offer 11.3 TFLOPs of computation,
have 11GB GDDR5X memory at 484GB/sec memory bandwidth, and cost only \textdollar 699);
and 3) accessiliby of state-of-the-art machine learning platforms.
A unique aspect of the recent deep learning revolution is that all major deep learning platforms, whether TensorFlow \cite{45381}, PyTorch \cite{paszke2017automatic}, Caffe \cite{Jia:2014:Caffe}, CNTK \cite{CNTK-Intro-YU+2014}, or MxNet \cite{DBLP:journals/corr/ChenLLLWWXXZZ15},
are all built as open-source software.
This enables rapid cross-pollination of ideas among researchers and engineers. It also allows big organization and individuals alike to use
state-of-the-art machine learning tools without incurring (traditionally significant) license fees, or the need to develop technology
from scratch.

The rapid development of AI systems has the opportunity to significantly impact our lives, aiding humans in making mission-critical or life-and-death decisions.
We believe that the task of skin lesion analysis toward melanoma detection organized by the International Skin Image Collaboration (ISIC)
is one such image classification task that will benefit from the use of state-of-the-art machine learning technology.
As discussed in \cite{DBLP:journals/corr/abs-1710-05006}, skin cancer is the most prevalent form of cancer in the United States.
Melanoma, the most dangerous form of skin cancer, leads to over 9,000 deaths a year. Most melanomas are first identified visually, but
unaided visual inspection only has a diagnostic accuracy of roughly 60\% \cite{kittler2002diagnostic}. Recently, medical professionals have used dermoscopy,
a new technology of visual inspection, that both magnifies the skin and eliminates surface reflection. With proper training, it is shown that a human expert
can achieve a diagnostic accurcy of 75\%-84\% \cite{kittler2002diagnostic,vestergaard2008dermoscopy}.
Moreover, a growing shortage in dermatologists per capita \cite{kimball2008us} leads to
interest in using AI techniques for automated assessment of dermoscopic images.

The rest of the paper will discuss our submission for the ISIC 2018 challenge on lesion segmentation and disease classification. We will briefly discuss our observation of the dataset and task in Section.~\ref{sec:dataset}. The methodology and results will be discussed in Section.~\ref{sec:method}.

\section{Dataset and Task}
\label{sec:dataset}




The HAM10000 Dataset \cite{tschandl2018ham10000} is used in our training pipeline. For task 1, Lesion Boundary Segmentation, we have 2594 images and corresponding ground truth response masks.
For task 3, Lesion Diagnosis, we have 10015 images used for training and validation.
There are 7 classes in the HAM10000 Dataset, including 1) melanoma, 2) Melanocytic nevus, 3) basal cell carcinoma, 4) Actinic keratosis/Bowen’s disease (intraepithelial carcinoma), 5) benign keratosis (solar lentigo/seborrheic keratosis/lichen planus-like keratosis), 6) dermatofibroma and 7) vascular lesion. The number of images in each category is extremely imbalanced. For example, the largest category, Melanocytic nevus, has 6705 images while the smallest category, dermatofibroma, only has 115 images.

\section{Method, Experiment Design and Results}
\label{sec:method}

When training human dermoscopic students, procedural algorithms, such as the ``3-point checklist'', the ``ABCDE rule'', and the ``7-point checklist'' were developed \cite{argenziano2003dermoscopy}. For example, the ``ABCDE'' rule suggests checking for suspicious moles using the following signs: 1) \underline{A}symmetry (one half of
the mole doesn't match the other), 2) \underline{B}order irregularity, 3) \underline{C}olor that is not uniform, 4) \underline{D}iameter greater than 6 mm (about the size of a pencil eraser), and 5) \underline{E}volving size, shape or color. There has been debate in the medical community about whether this procedural
guideline or relying on personal analysis is more beneficial. Similarly, when designing an AI system for dermoscopic image analysis, there can also be two different approaches.
First, there is the feature based approach, in which multiple sub-AI systems are developed, each of which will be trained to identify one feature (e.g., one AI sub-system each will be trained to measure \underline{A}symmetry, \underline{B}order irregularity, \underline{C}olor uniformity,  \underline{D}iameter and
\underline{E}volving size), and then an aggregated AI system will be used to combine the individual features and make disease diagnostics.
An alternative is the end-to-end approach, in which an AI system, in particular a deep convolutional neural network (CNN) will be trained directly from the image
datasets. We apply the end-to-end training approach, mainly because the current dataset lacks individual features (such as the ``ABCDE'' feature discussed above), and as a result, training multiple individual feature regression/classification subsystems may lead to added development cost. If the ISIC organizer later
providers metadata with individual features, especially features related to the \underline{E}volving size, shape, or color and
the \underline{D}iameter, both of which cannot be reliably estimated from the image by the end-to-end CNN approach, we will be interested to
investigate if the use of those individual features can improve disease detection accuracy.

In the ISIC 2018 challenge, it is required to only use images for lesion boundary segmentation and lesion diagnosis tasks. In Sec.~\ref{sec:method_task1}, we describe the method we used in Task 1 and in Sec.~\ref{sec:method_task3}, we describe the method we used in Task 3.

\subsection{Task 1: Lesion Boundary Segmentation}
\label{sec:method_task1}
We start this task by using the state-of-the-art object segmentation model Mask RCNN \cite{he2017mask}. We use ResNet50 as backbone network. The pre-trained weights on ImageNet dataset are used to initialize the backbone network. In this task, we do not separate the 7 categories in the training set. Instead, we treat all the lesion regions as one class, and add a background class during the training. We train the model on a single NVIDIA Titan X GPU. The network proposes 200 ROI candidates per image and classifies if each ROI contains a lesion region. We use a detection confidence threshold of 0.9 in our experiment to eliminate most of the false positive detections.  We also randomly select 20 images from the training set as validation set. Furthermore, we use the Jaccard index metric to measure our model. The model was trained for 40 epochs and we get 0.783 in our validation set.

We notice that the initialize backbone network weights was trained on the ImageNet dataset, which has 1000 classes of common visual objects. The ImageNet dataset is widely used in many computer vision applications, but its visual domain is quite different from the lesion image dataset. The first few layers of a CNN can be considered as feature extractor, and the ResNet50 model was trained on the ImageNet dataset, so the convolutional filters of the ResNet50 are tuned to extract features from the common objects that appeared in the ImageNet dataset. This causes the ResNet50 to sometimes be unable to locate the lesion region in the HAM10000 dataset. Thus, we create an improved solution: we first train the ResNet50 model using the data from the Lesion Diagnosis task, and after the model is well tuned and achieves a reasonable overall classification accuracy, we use the weights from this model to initialize the backbone network in our Mask RCNN model. With this method, we improve our validation performance to 0.818.

\subsection{Task 3: Lesion Diagnosis}
\label{sec:method_task3}

\begin{figure}
  \centering
  \includegraphics[width=0.6\textwidth]{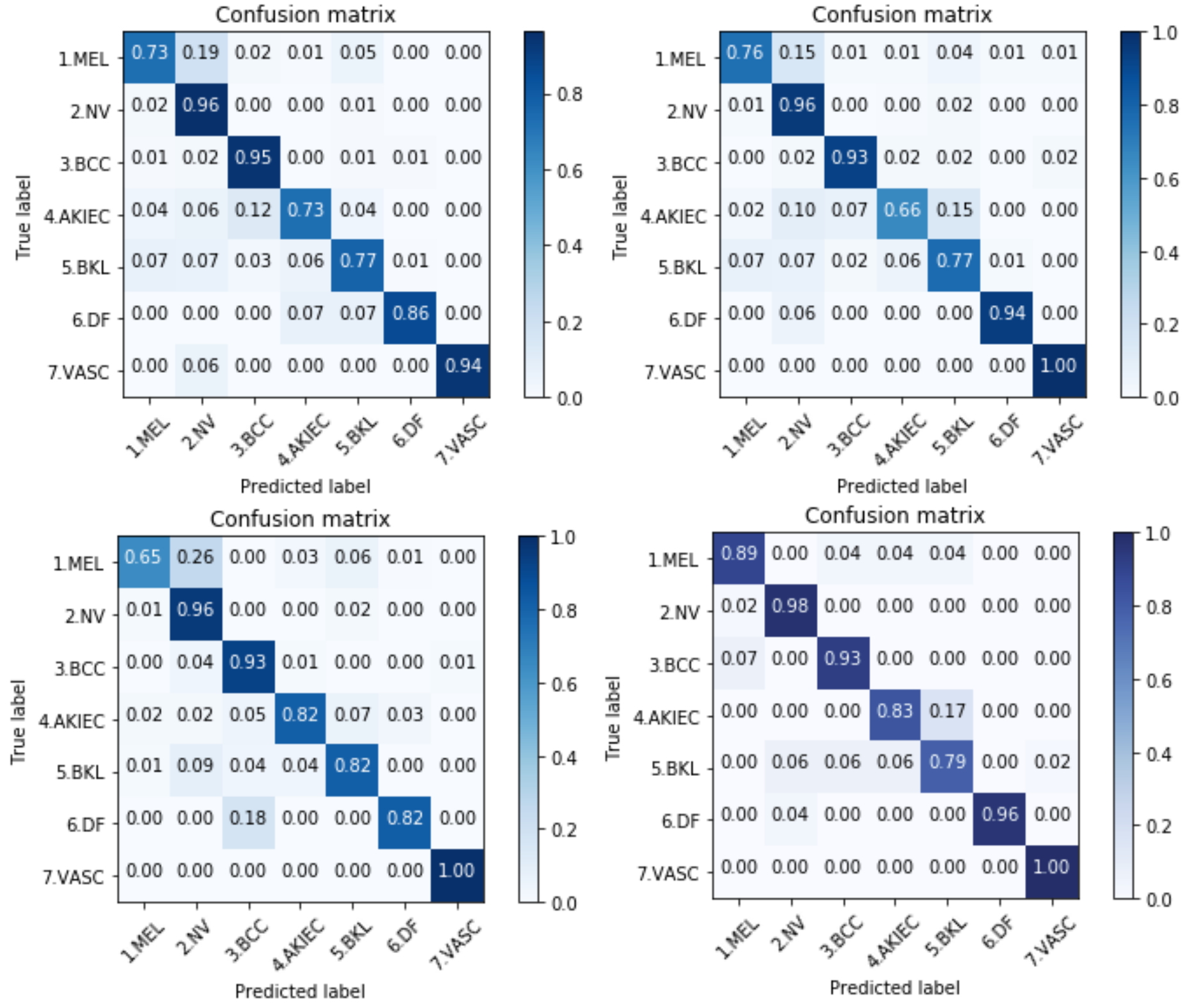}
  \caption[] {Confusion matrix from DenseNet201, ResNet152, Inception\_V4 and DenseNet201 with cropped images}
  \label{fig:results}
\end{figure}

We start our baseline experiment by training the multi-class classification model on a variety of CNN models, including ResNet, DenseNet and Inception. The dataset is split into 80\% training and 20\% validation set. We use a similar setup in each model. A global average pooling layer is added after the base network and followed by a fully connected layer for prediction. We use softmax as the activation function in the prediction layer and multi-class cross entropy as the loss function. Each model is trained for 300 epochs, and the loss and validation loss are converged after around 240 epochs.
As we mentioned in Section~\ref{sec:dataset}, the classes of the HAM10000 dataset are extremely imbalanced. We use class weights to balance the dataset. The higher weights will be given to the samples in the class with small size and the lower weights will be given to the samples in the class with large size.
We test our models in the validation set and the confusion matrix of each model is shown in Fig.~\ref{fig:results}.

\subsubsection{Training on Cropped Lesion Images}
We notice that we may improve the classification performance in the future by removing the background from the lesion images and letting the classifier only focus on the lesion region. Based on the segmentation model we get from Task 1, we can first perform lesion boundary segmentation on the dataset. The lesion region is cropped based on the segmentation results and used as new training/validation data. We achieve about 2\% improvement on normalized multi-class accuracy by using cropped images for training and validation. It is important to point out that this improvement highly depends on the performance of the lesion boundary segmentation model. The classification results may suffer if the performance of the segmentation model is not good enough.


\section{Conclusions}
This article summarized the methods we used in the ISIC 2018 challenge. We proposed several improvements to address the problem of Lesion Boundary Segmentation and Lesion Diagnosis.

\clearpage

\bibliographystyle{splncs}
\bibliography{egbib}

\end{document}